# Image Registration of Very Large Images via Genetic Programming

Sarit Chicotay
Dept. of Computer Science
Bar-Ilan University
Ramat-Gan, Israel
saritc@gmail.com

Eli (Omid) David
Dept. of Computer Science
Bar-Ilan University
Ramat-Gan, Israel
mail@elidavid.com

Nathan S. Netanyahu
Dept. of Computer Science
Bar-Ilan University
Ramat-Gan, Israel
nathan@cs.biu.ac.il

**Abstract**

*Image registration (IR) is a fundamental task in image processing for matching two or more images of the same scene taken at different times, from different viewpoints and/or by different sensors. Due to the enormous diversity of IR applications, automatic IR remains a challenging problem to this day. A wide range of techniques has been developed for various data types and problems. However, they might not handle effectively very large images, which give rise usually to more complex transformations, e.g., deformations and various other distortions.*

*In this paper we present a genetic programming (GP)-based approach for IR, which could offer a significant advantage in dealing with very large images, as it does not make any prior assumptions about the transformation model. Thus, by incorporating certain generic building blocks into the proposed GP framework, we hope to realize a large set of specialized transformations that should yield accurate registration of very large images.*

## 1. Introduction

Image registration (IR) is an important, significant component in many practical problems in diverse fields where multiple data sources are integrated/fused, in order to extract high-level information as to the contents of the given scene. A wide range of registration techniques has been developed over time, where typically, specific domain knowledge is taken into account and certain a priori assumptions are made, e.g., with respect to the transformation model used, specific bounds on its parameter values, etc.

In contrast, in this paper we present a *genetic programming* (GP)-based approach for IR. GP is part of a family of *evolutionary algorithms* (EAs) which are stochastic optimization methods whose goal is to find an "optimal" solution or a set of solutions with respect to certain objective(s). Exploiting this strength of GP for search and optimization problems [8-9], we use GP and image processing techniques to search efficiently for an "optimal transformation" with respect to a given similarity measure. The novel aspect of our proposed GP algorithm is that it does not make any prior assumptions about the transformation model. Therefore, incorporating various building blocks into the GP framework could yield potentially a large pool of transformations. The advantage of this approach is that it offers, in principle, much greater flexibility in the registration of very large images which give rise to various, relatively complex transformation types.

We present good results on some real datasets on small images and compare them to those obtained due to a recent method assuming a simple transformation model. We also present initial results on some real datasets containing larger images on simple transformations. From a pure GP perspective, other transformations, possibly more complex ones, could be searched for and discovered, as long as the GP is supported by the proper building blocks.

The paper is organized as follows. Section 2 reviews IR components that are essential for our solution. Section 3 presents a brief GP background and provides the motivation for our GP-based IR approach. Section 4 presents an overview of related evolutionary-based solutions for IR. Section 5 provides a detailed description of our suggested GP-based approach. In Section 6 we present our initial empirical results. Section 7 makes concluding remarks.

## 2. Image Registration

Image registration involves searching for a transformation that generates a maximal match in the overlap between the reference image and the transformed sensed image. It uses a similarity measure to assess the quality of a specific transformation.

In this work, we focus on one of the most common measures, *mutual information* (MI) [14],[18], inspired by information theory [17]. MI is a measure of statistical dependency between two datasets that has been applied in a robust and efficient manner to IR. Analogous to the Kullback-Leibler measure [15] for the distance between two distributions, MI of two images measures the degree of dependence between the gray values in the area of overlap, defined as,

$$(1) \quad I(A,B) = \sum_{a,b} p(a,b) \log \frac{p(a,b)}{p(a)p(b)}$$

The assumption is that maximal dependence between the gray values of the images achieved when they are correctly aligned, while misregistration results in a decrease of this measure. We considered close gray levels as having the same gray level for the purpose of calculating the MI.

The set of transformations (e.g., translation, rotation, scaling) that are available during the registration process define the search space. The search for an optimal transformation in terms of the similarity measure is carried out using a search strategy which governs how the search space is explored. The search strategy has a great impact on the efficiency of the image registration process.

In this paper we present an alternative GP-based search strategy. Since GP is known to be robust for a variety of search and optimization problems, it could also prove useful in the above scenario.

## 3. Genetic Programming

As noted, GP is part of evolutionary algorithms family inspired by the evolution in nature. It uses the principles of evolution and natural selection for optimization and search for the global min/max of a given objective function.

GP maintains a population of candidate solutions called chromosomes. The fundamental elements of an individual chromosome are its genes, which come together to form the tree-like program representing the chromosome consist of: (1) Functions, which in tree terminology, are nodes with children (the function's arguments) and (2) terminals which are leaves (i.e., nodes without branches). The functions can be basic mathematical functions (e.g., $+, -, *$) and possibly more complex ones. Terminals consist of the variables and constants of the program. See Figure 1.

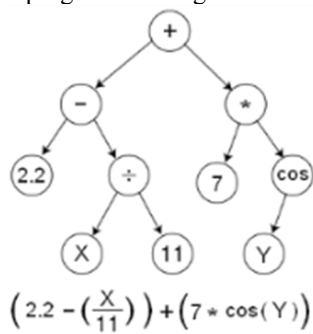

$$\left(2.2 - \left(\frac{X}{11}\right)\right) + (7 * \cos(Y))$$

**Figure 1: Example of a GP chromosome and the program it represents**

A basic GP algorithm consists of four iterative steps: (See Figure 2.)
1. Start with a randomly generated population of $N$ chromosomes.
2. Calculate the fitness value of each candidate in the population using a fitness function.
3. Repeat until $N$ offspring are created:
   3.1. Select a pair of chromosomes from the current population based on their fitness value (a chromosome can be selected more than once).
   3.2. Produce offspring using crossover between the two selected chromosomes and perform random mutation on those offspring.
4. Replace the population with the newly created one.

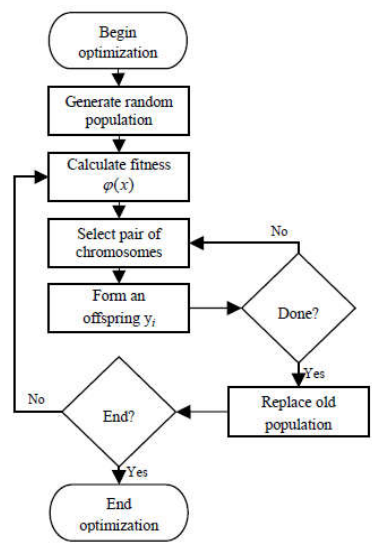

**Figure 2: Basic flow chart of GP**

The process of creating the offspring is called recombination and is combined of two steps: (1) Crossover, where the selected candidates are combined by swapping portions to produce hopefully better candidates with higher fitness for the next generation, and (2) mutation, which is performed on each of the new offspring with some low probability (mutation rate). Mutation generates variations in the population and helps avoid local minima. GP is designed to create and preserve a legal program structure during these steps.

The process continues iteratively until it satisfies certain termination criteria, such as finding an acceptable approximate solution, reaching a specific number of generations, or until the solution converges.

## 4. Related Work

Evolutionary algorithms (EAs) are stochastic optimization methods whose goal is to find an "optimal" solution or a set of solutions with respect to certain objective(s). EAs are considered robust methods, as one of their main advantages is to avoid local optima. These algorithms have been successful during the past decades in solving a variety of search and optimization problems including IR-related problems.

Most of the earliest evolutionary approaches for IR [19] used genetic algorithms (GAs) with proportionate selection and a binary representation. This approach has several limitations when solving optimization problems in the continuous domain due to the binary representation that is limited for each decision variable and selection method that have several drawbacks when compared to ordinal-based selection methods such as ranking or tournament selection [5]. Another limitation of the early approach that it only deals with translation and rotation, ignoring scaling and shearing.

Current EA studies for IR represent a GA-based approach [3],[11-12], which makes a prior assumption about the transformation model (e.g., similarity transformation). The optimal transformation is searched for in the space of transformation parameters.

Santamaria et al. [10] present a more detailed overview of evolutionary methods for image registration and some state-of-the-art methods.

In this work, we present a GP-based algorithm for IR. Unlike a GA approach, GP is not limited to a predefined number of transformation parameters. Therefore, incorporating various generic building blocks into the GP framework could yield potentially a diverse pool of transformation types, which should prove effective in the registration of very large images. A detailed description of this approach is given in the following section.

## 5. GP-Based Image Registration

This section describes our GP-based approach to optimize the IR similarity measures by utilizing tools common for IR.

We set the initial population to be twice as large as a predefined size in order to maintain enough diversity. For the next generation, elitism is used to retain the top 2% chromosomes from the current generation and the rest are taken from the new pool of chromosomes.

### 5.1. Chromosome Representation

For modeling an IR problem, the GP tree-like chromosome should represent a possible transformation. A transformation $T$ on a 2D image maps each pixel $p = (x, y)$ of the sensed image to a new pixel $p' = (x', y')$ in the coordinate system of the referenced image, i,e., $T(p) = p'$. More specifically, we obtain:

(2) $\quad x' = T_x(x, y), \ y' = T_y(x, y)$

Therefore, each chromosome in our solution will be comprised of two such GP trees, an $x$-tree and a $y$-tree, each representing the transformation on one of the coordinates. We incorporate into our GP framework a list of functions and terminals that are essential for composing a diverse set of global transformations. Inspired by the discussion on transformation models in [1-2], we define, in particular, the following "primitives".

#### 1.1.1 Function set

$Functions = \{+, -, *, /, cosine, sine, pow,$
$rotation_x(x, y), rotation_y(x, y), RBF(x, y), IRBF(x, y)\}$
where $x, y$ are the coordinates on which the transformation is performed.

$\{+, -, *, /, pow\}$ – have arity of two, i.e., 2 child nodes.
$\{cosine, sine, rotation_x, rotation_y\}$ - 1 child.
$\{RBF, IRBF\}$ – 3 child nodes.

$rotation_x$, $rotation_y$ are functions of $x, y$ and a child node; they describe a 2D rotation of an image around its center, i.e.,
(3) $rotation_x(x, y, child) = (x - width/2)$
$\cos(child) - (y - height/2)\sin(child) + width/2$
(4) $rotation_y(x, y, child) = (y - height/2)$
$\cos(child) + (x - width/2)\sin(child) + height/2$

$RBF = Radial\ Basis\ Function$, is a function of $x, y$ and its three child nodes, defined as follows:
(5) $RBF(x, y, child_1, child_2, child_3) =$
$\sqrt{(x - child_1)^2 + (y - child_2) + child_3{}^2}$

$IRBF = Inverse\ RBF$, is a function of $x, y$ and its three child nodes, defined as follows:
(6) $IRBF(x, y, child_1, child_2, child_3) =$
$\left(\sqrt{(x - child_1)^2 + (y - child_2) + child_3{}^2}\right)^{-1}$

Although we have included some special functions as prospective building blocks of the chromosome trees, the GP algorithm may not necessarily use them "as-is". Despite this, we include these specific elements in order to assist in the convergence of the GP algorithm, should their explicit use become relevant.

#### 1.1.2 Terminal set

$Terminals = \{constant, variable, e\}$

A $constant$ is some random number selected uniformly between $\pm max\{width, height\}$, where $width, height$ are the dimensions of the sensed image.

A $variable$ can be one of the coordinates, $x$ or $y$.

$e$ is the base of the natural logarithm $\approx 2.71828$

### 5.2. Crossover Operator

Crossover is carried out separately on the $x$-tree and $y$-tree of the two chromosomes. We used a simple one-point crossover, i.e., we select a crossover point in each of the two matching trees and then swap the corresponding sub-trees. We also allow crossover to take place between an $x$-tree and a $y$-tree with a low probability of ~0.2. This enhances the possibility of maintaining consistent components by both transformations (e.g., rotation, scale). The resulting chromosomes are added to the new pool.

### 5.3. Mutation Operator

We have used the following two mutation variants:
1. Mutation of a randomly selected node in the tree, by replacing its sub-tree with a random generated sub-tree (with limited depth.)
2. Mutation of each node with some low probability by replacing it with a random node (function/terminal) with the same arity (i.e., same number of branches.)

The chromosome created by recombination is mutated with probability 0.3, i.e., each of the above two variants is invoked with probability of 0.15.

When the algorithm reaches a steady fitness over a predefined number of generations, the mutation rate is incremented by 0.02. If the fitness continues to improve the mutation rate is reset to its original value.

### 5.4. Fitness Function

The IR goal is to optimize the similarity measure(s) between the transformed sensed image and the referenced image. We used MI as mentioned in Section 2.

In order for this function to be relevant for the registration of large images, we used the following sampling strategy. For each fitness evaluation, we select randomly a predefined number of pixels in the sensed image (typically 0.5% of the pixels.) The fitness calculation is performed on these pixels only. The overlap percentage between the images is also calculated only with respect to these randomly selected pixels, a chromosome associated with an overlap that is smaller than a predefined threshold is discarded. This strategy speeds up computation and has been proven useful in our initial tests.

### 5.5. Selection Operator

We used rank-based selection which is considered more robust than other selection methods [7],[16]. The chromosomes are ordered by their fitness values due to the MI measure. Then, each chromosome is given a rank according to its position (the lower the better). Two chromosomes are selected for recombination based on their calculated ranking. The probability for an individual position $pos$ to be selected within a population of size $M$ is given by:

(7) $$\frac{(M - pos + 1)}{\sum_{i=1}^{M} i}$$

### 5.6. Stopping Criteria

As with most stochastic optimization algorithms, we have no clear way of knowing when to stop the search and accept the current best solution as the (near) optimal solution. We stop the algorithm when the solution is not "significantly" updated for a predefined number of generations. The chromosome with rank 1 at this stage is chosen as the suggested solution by the GP.

## 6. Empirical Results

The following parameter values were used consistently by our GP-based IR algorithm for all evaluations:

| Population size | 150 |
| Crossover probability | 0.9 |
| Mutation rate | 0.3 |
| Elitism | 3/150 |
| Initial tree max. height | 6 |
| Mutation max. height | 3 |

Table 1: Parameter values of the GP algorithm

The GP approach was initially tested on the 256x256 images described in [13]. We used several sub-images as reference images ("chips") and registered to them overlapping sub-images sensed at different times. Several examples are shown in Figure 3.

In order to assess the correctness of the final transformation, we selected manually $N$ points (typically about 10 points) in the reference image $\{(x_i, y_i)\}_{i=1,\ldots,N}$ and their corresponding points in the input image $\{(x'_i, y'_i)\}_{i=1,\ldots,N}$, respectively. We consider these sets as the ground truth (i.e., most accurate transformation).

Next we apply the transformation to each point in $\{(x'_i, y'_i)\}_{i=1,\ldots,N}$, getting $\{(\tilde{x}_i, \tilde{y}_i)\}_{i=1,\ldots,N}$, and compute the root mean square error (RMSE) defined as:

(8) $$RMSE = \sqrt{\frac{1}{N}\sum_{i=1}^{N}(x_i - \tilde{x}_i)^2 + (y_i - \tilde{y}_i)^2}$$

We achieved an RMSE value of ~1 pixel on the examples shown in Figure 3, which is considered a successful registration. We compared also our results to another method tested on the same datasets [4]. The results obtained by both methods are comparable. Nevertheless, the quality of the results due to our GP based method should not be underestimated as no assumption was made about the transformation model in order to initiate or bound the registration process.

Table 2 presents some of our results. Notice that the resulting transformations do not necessarily fit a familiar structure of a transformation combined of translation, rotation, and scaling. The GP scheme produces various transformations as part of evolving improved transformations, in an attempt to meet the optimal similarity measure. As with most IR applications, the objective is to align the images so that additional high-level operations could be performed. Thus, the characteristics of the transform are not that important, as long as the alignment is achieved with high level of accuracy.

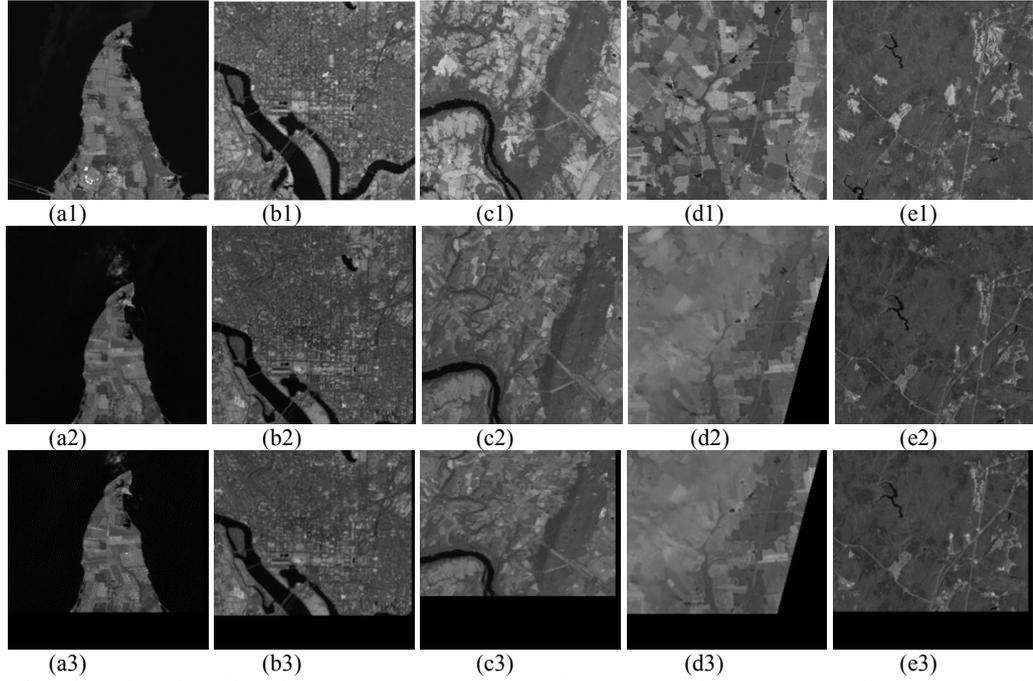

**Figure 3:** (a1), (b1), (c1), (d1), (e1): 256x256 reference images over the Washington, DC area, (a2), (b2), (c2), (d2), (e2): corresponding 256x256 sensed images, and (a3), (b3), (c3), (d3), (e3): registration results.

Motivated by the results on the set of 256x256 images, we have been exploring the validity of our approach on several, much larger space/airborne images. Our initial tests were performed on a Landsat 7 scene over Colorado (from NASA's IMAGESEER database). The scene is of size 5351x5051, (~27M pixels) and contains ~5% cloud coverage. We created two semi-synthetic image pairs. The first pair was created by applying only a translation with the ground truth transformation: $T_x(x,y): x - 50$, $T_y(x,y): y - 100$. We obtained, in this case, a successful registration with ~1 pixel RMSE. The second pair was created by applying both a rotation and translation with the ground truth transformation: (angles are in radians):
$T_x(x,y): rotate_x(0.2618) + 15$,
$T_y(x,y): rotate_y(0.2618) - 165$.

Figure 4 shows the registration result obtained with an RMSE of ~6 pixels. The transformation obtained by the GP scheme in this case (angles are in radians) is:
$T_x(x,y): \sin(3.714) + (12.24 + rotate_x(((((-1953.79) + 1995.74) * \sin(-6.05)) - (-5048.61))))$
$T_y(x,y): ((rotate_y(\cos(550.67 - (-420.77) + (-69.61)))/69.58) + (((rotate_y(((((9.78 + 550.67) + ((\cos(-94.34) + (-98.93)) + (-98.93))/((-340.51)/1937.69) * (-13.18) ))))/(9.78 + rotate_x(((((-1953.79) + 1995.74) * \sin(-81.45)) - (-5048.61))) ) + ((rotate_y(\cos(((((2185.03/(-89.18)) - (-1055.97)) - 2301.13) + (\cos(926.18) + (-98.93))) - (-98.93)))) ) + (-98.93)) + (-98.93))))$

| | SIFT-based IR [4] | | | | | GP-Based IR | | |
|---|---|---|---|---|---|---|---|---|
| Image pair | s | θ [deg] | $t_x$ [pix] | $t_y$ [pix] | RMSE [pix] | $T_x(x,y)$ | $T_y(x,y)$ | RMSE [pix] |
| (a1)-(a2) | 1.001 | -0.24 | -5.76 | -46.39 | 0.74 | $((x-\cos(358.26))-4.23)$ | $((2.7+(y+\sin(128.33)))-48.99)$ | 0.85 |
| (b1)-(b2) | 1.006 | -0.27 | -4.75 | -47.61 | 0.86 | $((\cos(202.16)+\cos(202.16))+\sin(242.67)+((x+\cos(183.68))+\sin(278.16))))$ | $((((y-\sin(8.74))-\cos(66.27))-(43.03-\sin(209.9))-\sin(209.9)))+\sin(254.8))$ | 1.04 |
| (c1)-(c2) | 1.003 | -0.47 | -4.55 | -48.73 | 0.63 | $((\sin(310.19)*4.17)+x$ | $((\sin(2.32)+\cos(267.27))+((8.64+y)+(59.4*\sin(284.74))))$ | 1.3 |
| (d1)-(d2) | 1.038 | 2.38 | -6.36 | -46.18 | 0.76 | $((0.47-5.11)+x)$ | $(\cos(24.8)+(y-45.05))$ | 1.3 |
| (e1)-(e2) | 1.002 | -0.006 | -5.4 | -47.7 | 0.6 | $((\sin(319.21)+(\sin(319.21)+\cos(188.9)))+(\sin(309.02)+((x+\sin(208.03))+\sin(319.21))))$ | $(((y-\sin(39.8))-46.41)-\cos(92.43))$ | 1.22 |

**Table 2:** Comparison of registration results due to our GP-based algorithm and a SIFT-based method [4] (angles are in degrees)

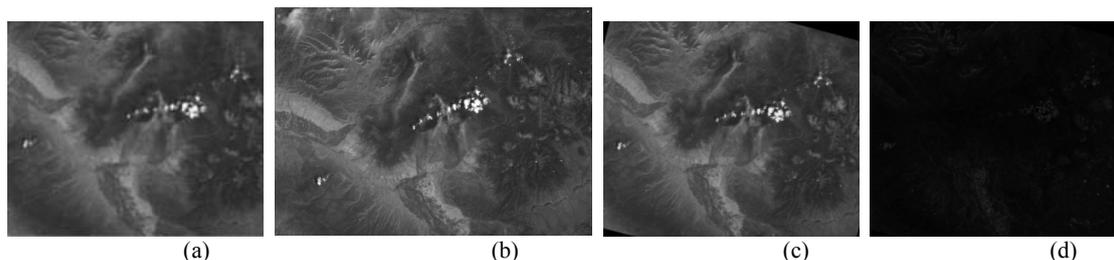

| (a) | (b) | (c) | (d) |

**Figure 4:** (a)-(b) Two Landsat 7 images to be aligned: (a) reference image (5351x5051 pixels), (b) semi-synthetic sensed image (6151x5151 pixels), (c) registration of sensed image to reference image, and (d) difference between reference image (a) and transformed sensed image (c) in overlap area.

## 7. Conclusions

We presented a GP-based image registration algorithm, whose main novelty over existing IR techniques is that it attempts to provide a robust and automatic solution without assuming any transformation model. The initial encouraging results demonstrate the potential of the powerful, evolutionary GP approach, as far as its application to IR. The accuracy of the results was assessed using RMSE, and was found comparable to another IR method presented recently.

The initial tests suggest that this approach can be potentially useful when considering very large images that give rise to more complex transformations. This would be further investigated as part of future research.

We intend to test extensively this scheme on additional, challenging datasets containing deformations and other distortions. Also, we will incorporate more of the common similarity metrics used in IR in addition to the MI measure used in this work (e.g., PHD, cross-correlation, sum of squared intensity) and consider additional functional building blocks for our GP-based scheme.

In view of our initial results, we believe there is indeed promise for such future directions.